\documentclass[10pt,twocolumn,letterpaper]{article}

\usepackage{cvpr}      
\usepackage{url}
\usepackage{natbib}
\usepackage{amsmath}
\usepackage{caption}
\usepackage{booktabs}      
\usepackage{multirow}
\usepackage{graphicx}
\usepackage{float}
\usepackage{pifont}
\usepackage{url}            
\usepackage{booktabs}       
\usepackage{amsfonts}       
\usepackage{nicefrac}       
\usepackage{microtype}      
\usepackage{amsmath, amssymb, amsthm}
\usepackage{colortbl}
\usepackage{pifont} 
\usepackage{multirow}
\usepackage{multicol}
\usepackage{makecell}
\usepackage{graphicx}
\usepackage{diagbox}
\usepackage{subcaption}
\usepackage[utf8]{inputenc}
\usepackage{svg}
\usepackage{comment}

\usepackage{float}
\usepackage{enumitem}
\usepackage{caption}
\usepackage{graphicx}
\usepackage{caption}
\usepackage{booktabs}
\usepackage{graphicx}    
\usepackage{caption}     
\usepackage{subcaption}  
\definecolor{cvprblue}{rgb}{0.21,0.49,0.74}
\usepackage[pagebackref,breaklinks,colorlinks,allcolors=cvprblue]{hyperref}

\def\paperID{} 
\def\confName{CVPR}
\def\confYear{2026}

\title{From Sketch to Fresco: Efficient Diffusion Transformer with Progressive Resolution}

\author{ \bf
Shikang Zheng\textsuperscript{1,2},
Guantao Chen\textsuperscript{1},
Lixuan He\textsuperscript{3},
Jiacheng Liu\textsuperscript{1},
Yuqi Lin\textsuperscript{1},\\ \bf
Chang Zou\textsuperscript{1},
Linfeng Zhang\textsuperscript{1,\dag}
\\[0.3em]
\textsuperscript{1}Shanghai Jiao Tong University,
\textsuperscript{2}South China University of Technology,
\textsuperscript{3}Tsinghua University\\
}

\begin{document}
\maketitle

\renewcommand{\thefootnote}{\fnsymbol{footnote}}
\footnotetext[2]{Corresponding author.}
\renewcommand{\thefootnote}{\arabic{footnote}}

\begin{abstract}
Diffusion Transformers achieve impressive generative quality but remain computationally expensive due to iterative sampling.  Recently, dynamic resolution sampling has emerged as a promising acceleration technique by reducing the resolution of early sampling steps. However, existing methods rely on heuristic re-noising at every resolution transition, injecting noise that breaks cross-stage consistency and forces the model to relearn global structure. In addition, these methods indiscriminately upsample the entire latent space at once without checking which regions have actually converged, causing accumulated errors, and visible artifacts. Therefore, we propose \textbf{Fresco}, a dynamic resolution framework that unifies re-noise and global structure across stages with progressive upsampling, preserving both the efficiency of low-resolution drafting and the fidelity of high-resolution refinement, with all stages aligned toward the same final target. Fresco achieves near-lossless acceleration across diverse domains and models, including 10$\times$ speedup on FLUX, and 5$\times$ on HunyuanVideo, while remaining orthogonal to distillation, quantization and feature caching, reaching 22$\times$ speedup when combined with distilled models. Our code is in supplementary material and will be released on Github.
\end{abstract}

\section{Introduction}

\begin{figure}[htbp]
  \centering
  \includegraphics[trim=200 90 215 80, clip,width=1\linewidth]{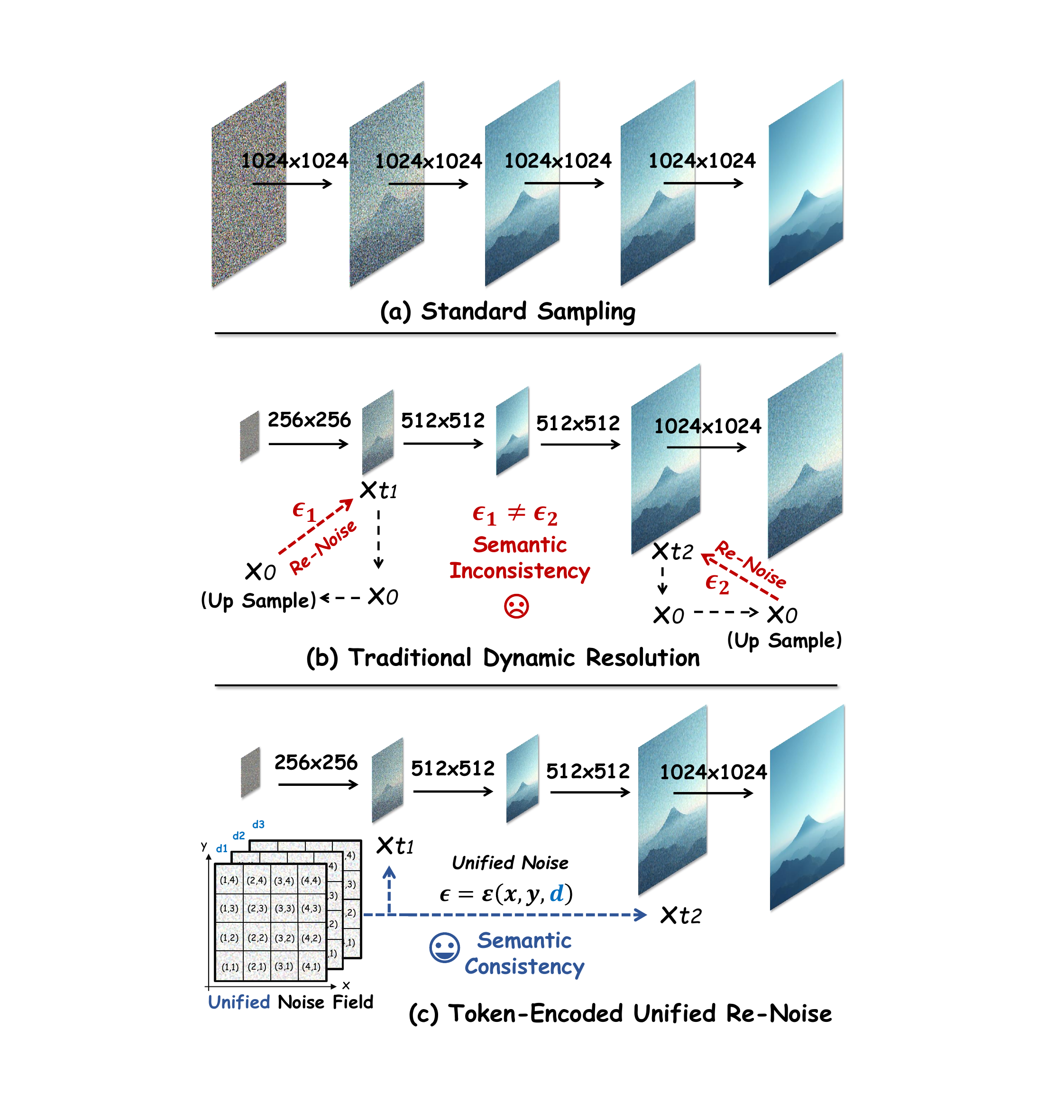}
  \caption{\textbf{Overview of re-noising strategies.} (a) Standard Sampling: one initial noise with no re-noise during process. (b) Traditional Dynamic Resolution: inject stage-specific noise independently at every resolution change, disrupting semantic and reset denoising trajectory, causing aliasing and artifacts. (c) Unified re-noise (ours): all stages query the same noise field, ensuring stable refinement and clean results.}
  \vspace{-5mm}
  \label{fig:intro}
\end{figure}

Diffusion Transformers (DiTs) have recently demonstrated remarkable performance on image and video generation tasks, achieving state-of-the-art fidelity and generalization. However, the inherent iterative nature of diffusion sampling poses a fundamental challenge: each output sample requires dozens of forward passes through a large transformer backbone. This leads to prohibitive computational costs, making real-time or resource-constrained deployment infeasible and fueling the search for more efficient inference techniques.

To address this challenge, two primary acceleration directions have emerged: reducing the total number of sampling steps through higher-order solvers~\citep{lu2022dpm, zheng2023dpmsolvervF}, distillation~\citep{salimans2022progressive}, and consistency training~\citep{song2023consistency}, while another line of work focuses on reducing the computation per step by optimizing model operations, such as sparse attention~\citep{child2019generating, zaheer2020bigbird} and feature caching~\citep{ma2024deepcache, liu2025reusingforecastingacceleratingdiffusion}. Recently, dynamic resolution sampling has emerged as a promising technique: by performing early sampling steps at lower resolutions, it significantly reduces the computation cost in the initial stages~\citep{jeong2025upsamplemattersregionadaptivelatent,tian2025trainingfreediffusionaccelerationbottleneck}. Despite progress, it introduces new challenges.

Dynamic resolution approaches often involve multiple resolution transformations across sampling stages, where each transition to a new resolution requires a re-noise operation that injects fresh noise to align with the forward diffusion process. However, current methods typically rely on intricate re-noise schedules and stage-specific heuristics as shown in Fig.~\ref{fig:intro}(b). This not only complicates implementation and tuning, but also limits the achievable acceleration.

Beyond complexity, a more fundamental issue lies in the lack of cross-stage consistency introduced by these re-noise steps. After each resolution change, a newly sampled noise field is added independently, decoupled from the denoising trajectory at previous stages. While this preserves randomness, it acts as a destructive reset: the model's generation path, previously converging toward a meaningful low-resolution draft, is abruptly diverted by a fresh perturbation at higher resolution. As a result, the model is forced to relearn global semantics instead of refining details for existing content, leading to unstable textures, broken geometry, and limited acceleration. This trajectory misalignment undermines the core advantage of low-resolution generation, where global structure and semantic intent can be captured rapidly with minimal computational cost. 


Another limitation of previous methods is their indiscriminate one-step upsampling of the entire latent space, performed without any principled criterion to assess the convergence state of individual tokens at low resolution. Consequently, spatial regions that have not yet stabilized semantically are upsampled prematurely, disrupting the natural progression of the diffusion process, which often leads to visible aliasing, ringing, and fragmented geometry.


Therefore, we introduce \textbf{Fresco}, a training-free progressive resolution framework that unifies global structure and semantics across the entire process. As shown in Fig.~\ref{fig:intro}(c), Fresco builds a token-encoded unified noise field, where each token is assigned a fixed noise vector shared across all stages, ensuring consistent stochastic evolution. Meanwhile, tokens are progressively upsampled based on their variance: those with lower variance, indicating stable structure and semantics, are promoted early to higher resolution for detailed refinement. Others remain at low resolution until their representations stabilize. This mechanism combines the rapid convergence of low-resolution generation draft and the detailed refinement of high-resolution sampling steps, with both stages aiming for the same final target. Remarkably, our approach achieves high-fidelity generation with a 10$\times$ speedup on FLUX and 5$\times$ on HunyuanVideo without extra training. Moreover, it remains orthogonal to other acceleration methods such as distillation, quantization and feature caching, reaching up to 22$\times$ acceleration when combined with distillation. In summary, our main contributions are:
\begin{itemize}
    \item \textbf{Unified noise field.}
     We construct a globally consistent noise representation by assigning each token a fixed noise vector shared across all stages, which maintains structure and coherent semantic evolution.
    
    \item \textbf{Adaptive progressive upsampling.} We introduce a upsampling mechanism that promotes tokens to higher resolutions based on their variance progressively, coupling fast convergence with detailed refinement.
    
    \item \textbf{Outstanding performance.} We evaluate Fresco across diverse models and tasks at multiple resolutions. In all settings, Fresco delivers state-of-the-art performance. Moreover, Fresco integrates well with other acceleration, providing additional performance gains across all settings without retraining.
\end{itemize}

\section{Related Work}

Diffusion models~\citep{sohl2015deep,ho2020DDPM} have demonstrated remarkable capabilities in image and video generation. While early implementations commonly relied on U-Net architectures~\citep{ronneberger2015unet}, their scalability constraints were later mitigated by Diffusion Transformers~\citep{peebles2023dit}, which enabled significant improvements in quality and resolution across multiple modalities~\citep{chen2023pixartalpha,chen2024pixartsigma,opensora,yang2025cogvideox}. Despite these advances, the core limitation of diffusion models remains their iterative sampling process, which introduces substantial computational overhead during inference. This has motivated two key research directions: minimizing the number of sampling steps, and reducing the per-step computational cost. Beyond raw efficiency, an ongoing challenge lies in preserving generative fidelity and stability, particularly under aggressive acceleration.

\begin{figure*}[t]
  \centering
  \includegraphics[trim=20 40 10 160, clip,width=1\linewidth]{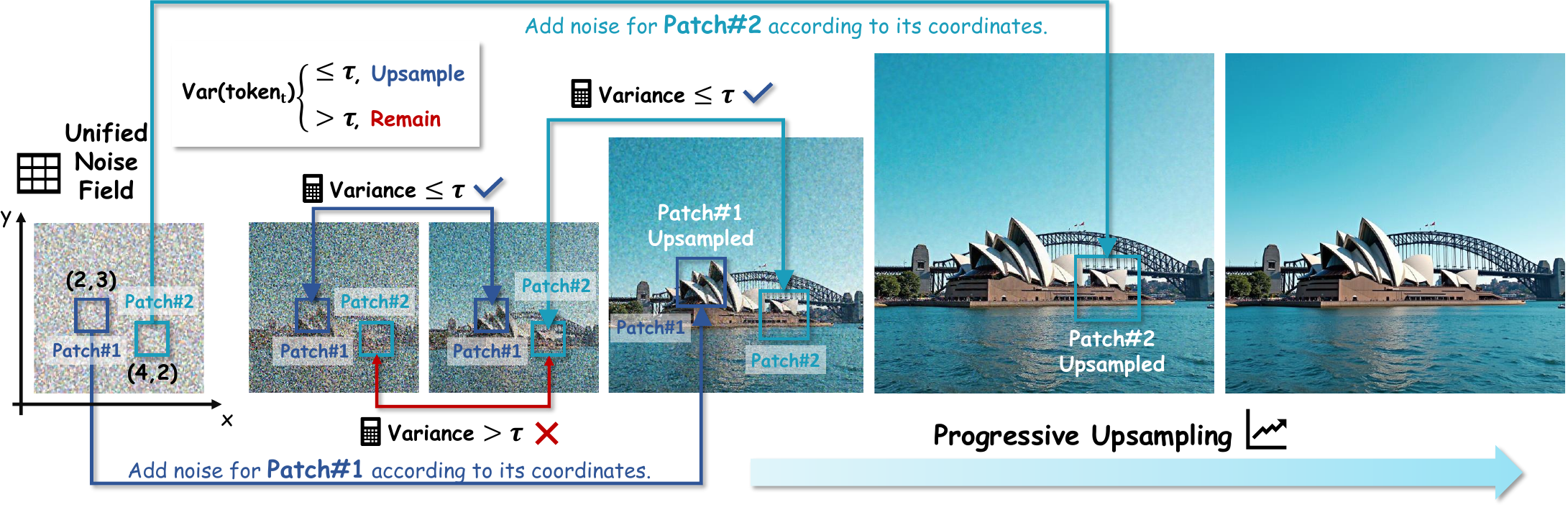}
  \caption{\textbf{Overview of the Fresco framework.} Fresco starts sampling at a reduced resolution while assigning each token a fixed coordinate-bound noise vector from a unified noise field. During generation, Fresco tracks each token’s temporal variance: tokens with small variance, indicating stable semantics, are upsampled for high-resolution refinement, whereas unstable tokens remain at low resolution for further denoising. This unified-noise, variance-guided process enables smooth and efficient coarse-to-fine generation.}
  \label{fig:framework}
\end{figure*}

\subsection{Temporal acceleration.}
A major line of research focuses on reducing the number of sampling timesteps to accelerate inference.
DDIM~\citep{songDDIM} first introduced a deterministic sampler enabling faster generation with minimal perceptual degradation.
Subsequent advancements leverage higher-order numerical solvers such as DPM-Solver and its improved variants~\citep{lu2022dpm, lu2022dpm++, zheng2023dpmsolvervF}, which offer better trade-offs between quality and efficiency by minimizing local truncation error.
Other approaches like Rectified Flow~\citep{refitiedflow} design alternative training objectives that result in shorter data transport trajectories, while progressive distillation~\citep{salimans2022progressive} compresses long denoising chains into shallow student models. Consistency models~\citep{song2023consistency} further reduce inference steps by directly learning a fixed-point mapping from noise to signal. Another complementary direction is training-free feature caching, which aims to reduce redundant computation by reusing previously computed hidden states or attention maps across timesteps.

\subsection{Spatial acceleration}

Sparse attention methods, such as fixed-window, strided, and adaptive patterns~\citep{beltagy2020longformer, zaheer2020bigbird, child2019generating}, aim to reduce the quadratic cost of self-attention by limiting token interactions. However, despite their elegance, the actual speedups achieved by sparse attention are often limited in practice. More substantial gains come from reducing spatial resolution during generation. Cascade diffusion frameworks~\citep{luo2023videofusion, ho2022cascaded, li2022srdiff, saharia2022image} follow a coarse-to-fine strategy, generating low-resolution outputs and refining them via learned upsampling.
While effective, these approaches typically require retraining or auxiliary modules. Recently, to improve flexibility, training-free dynamic resolution sampling methods such as bottleneck sampling~\citep{tian2025trainingfreediffusionaccelerationbottleneck} have been proposed, which reduces resolution in certain stages during sampling. However, these approaches heavily rely on intricate re-noising schedules and complicated parameter tuning, making it difficult to apply robustly in practice and limiting their achievable acceleration ratio.

\section{Method}

\subsection{Preliminary}

\noindent\textbf{Diffusion Models.} Diffusion models generate by simulating a forward–reverse stochastic process. The forward process gradually corrupts a data sample \(x_0\) with Gaussian noise:
\begin{equation}
    x_t = \sqrt{\alpha_t} x_0 + \sqrt{1 - \alpha_t} \, \epsilon_t, \quad \epsilon_t \sim \mathcal{N}(0, \mathbf{I}),
\end{equation}
where \(\alpha_t\) is a decreasing noise schedule. After \(T\) steps, the sample approaches pure noise. The reverse process learns to recover \(x_0\) by predicting the noise component \(\epsilon_\theta(x_t, t)\), commonly parameterized as:
\begin{equation}
    x_{t-1} = \frac{1}{\sqrt{\alpha_t}} \left( x_t - \frac{1 - \alpha_t}{\sqrt{1 - \bar{\alpha}_t}} \, \epsilon_\theta(x_t, t) \right) + \sigma_t \, \epsilon,
\end{equation}
where \(\bar{\alpha}_t = \prod_{s=1}^t \alpha_s\), and \(\sigma_t\) controls residual noise.

\noindent\textbf{Dynamic Resolution.} Dynamic resolution methods accelerate diffusion by reducing the spatial resolution of latent features during early timesteps, where fine details are less important. 
Given a latent feature map \(\mathbf{x}_t \in \mathbb{R}^{H \times W \times C}\), it is downsampled and then upsampled using:
\[
\hat{\mathbf{x}}_t = \mathcal{U}_s\big(f(\mathcal{D}_s(\mathbf{x}_t))\big),
\]
where \(\mathcal{D}_s\) and \(\mathcal{U}_s\) denote downsampling and upsampling operators, respectively, and \(f(\cdot)\) represents the denoising process at the reduced resolution. When the latent representation is rescaled, the noise distribution no longer matches the expected variance of the diffusion trajectory at the new resolution. Thus, an additional noise term must be injected to restore the correct stochastic balance, ensuring that the new latent \(\hat{\mathbf{x}}_t\) remains aligned with the diffusion prior. However, existing re-noising strategies typically rely on rescheduling, resetting the noise level to a few steps earlier and regenerating at higher resolution. Such repeated rescheduling introduces extra complexity, depends heavily on limited historical information, and often requires extensive manual tuning.

\subsection{Overview of Fresco}

Fresco is a training-free progressive resolution framework designed to connect the efficiency of low-resolution generation with the fidelity of high-resolution synthesis. Instead of empirically and independently re-noising at each resolution change, Fresco assigns every token a fixed coordinate-bound noise vector, forming a unified noise field that remains consistent across all stages. The sampling process begins with an initial latent $\mathbf{z}_0$ at reduced resolution, allowing early steps to capture the global structure efficiently. As sampling proceeds, Fresco determines the upsampling schedule for each token based on its temporal variance. Once a token's variance becomes sufficiently small, indicating stable semantics at the current scale, it is selectively upsampled for high-resolution refinement, while tokens with higher variance continue evolving at lower resolution.

\subsubsection{Unified Noise Field}

To simplify the re-noise process, reduce semantic drift across stages, and minimize heuristic tuning, we introduce a \emph{Token-Encoded Unified Noise Field}, which provides a single deterministic reference shared consistently across all stages. We define a global field where each latent token is assigned a fixed Gaussian vector according to its spatial coordinate and feature index. Formally,  
\begin{equation}
\epsilon_{y,x,d} = \mathcal{N}(0,1;\,\text{seed}=h(y,x,d)),
\end{equation}
where $h(\cdot)$ is a hash-based function ensuring that the same token always receives the same noise value across the entire sampling trajectory. During each resolution transition, the latent state is updated using this unified field:
\begin{equation}
\mathbf{z}^{(s+1)} = \beta_s\,\mathbf{z}^{(s)} + \alpha_s\,\epsilon_{y,x,d},
\end{equation}
with coefficients $(\alpha_s,\beta_s)$ controlling the stochastic contribution. While upsampling, new token coordinates are derived from their parent locations, and their noise values are directly queried from the same field according to their new coordinates.  
This coordinate-consistent construction guarantees that stochastic evolution remains continuous across spatial scales, avoiding destructive resets while preserving semantic alignment throughout the diffusion process.

\noindent\textbf{Proposition 1 (Unified re-noising yields smaller trajectory error; proof in A.1).}
\textit{
Let $\widehat X_e$ denote the state obtained by a unified re-noise update and $\widetilde X_e$ denote the state obtained by an independent-stage re-noise update at the same transition. The deviation from the target $X(t_e)$ satisfies the strict ordering:
\begin{equation}
\mathbb{E}\!\left[\big\|\widehat X_e - X(t_e)\big\|^2\right]
\;\le\;
\mathbb{E}\!\left[\big\|\widetilde X_e - X(t_e)\big\|^2\right],
\end{equation}
and, in particular, the independent-stage update incurs an unavoidable lower bound:
\begin{equation}
\mathbb{E}\!\left[\big\|\widetilde X_e - X(t_e)\big\|^2\right]
\;\ge\; b^2\,d,
\end{equation}
with $b$ the injected noise strength and $d$ the feature dimension. Therefore, re-noising with a shared noise realization is effectively a time reparameterization along the same generation path, yielding negligible drift, whereas injecting an independent new noise at each stage introduces an irreducible extra deviation whose expected squared magnitude grows proportionally to $b^2 d$.}

\begin{table*}[htbp]
\centering
\caption{\textbf{Quantitative comparison of text-to-image generation} on FLUX.1-dev.}
\setlength\tabcolsep{7pt} 
\small
\resizebox{\textwidth}{!}{
\begin{tabular}{l | c | c  c | c  c | c | c }
    \toprule
    \rowcolor{white}
    \multirow{2}{*}{\centering \bf Method} & \multirow{2}{*}{\centering \bf NFE} & \multicolumn{4}{c|}{\bf Acceleration} & \multirow{2}{*} {\bf Image Reward $\uparrow$} & \multirow{2}{*}{\bf CLIP Score $\uparrow$}\rule{0pt}{2ex} \\
    \cline{3-6}
    & & {\bf Latency(s) $\downarrow$} & {\bf Speed $\uparrow$} & {\bf FLOPs(T) $\downarrow$}  & {\bf Speed $\uparrow$} &  & \\

    \midrule
    {FLUX.1-dev} & 50 & 25.82 & 1.00$\times$ & 3719.50 & 1.00$\times$ & 
    0.9736 \textcolor{gray!70}{\scriptsize (+0.00\%)} & 
    32.404 \textcolor{gray!70}{\scriptsize (+0.00\%)} \\ 
    \midrule

    $60\%$ steps & 30 & 16.70 & 1.55$\times$ & 2231.70 & 1.67$\times$ &
    0.9663 \textcolor{gray!70}{\scriptsize (-0.75\%)} &
    32.312 \textcolor{gray!70}{\scriptsize (-0.28\%)} \\

    SpargeAttention & 50 & 15.46 & 1.67$\times$ & 2150.02 & 1.73$\times$ &
    0.9812 \textcolor{gray!70}{\scriptsize (+0.78\%)} &
    32.391 \textcolor{gray!70}{\scriptsize (-0.04\%)} \\
    
    TeaCache $({l}=0.25)$  & 50 & 14.10 & 1.83$\times$ & 1937.24 & 1.92$\times$ &
    0.9449 \textcolor{gray!70}{\scriptsize (-2.95\%)} &
    32.247 \textcolor{gray!70}{\scriptsize (-0.48\%)} \\

    TaylorSeer $(\mathcal{N}=3)$ & 50 & 9.89 & 2.61$\times$ & 1320.07 & 2.82$\times$ &
    0.9889 \textcolor{gray!70}{\scriptsize (+1.57\%)} &
    32.413 \textcolor{gray!70}{\scriptsize (+0.03\%)} \\

    Bottleneck Sampling & 30 & 11.32 & 2.28$\times$ & 1582.77 & 2.35$\times$ &
    0.9739 \textcolor{gray!70}{\scriptsize (+0.03\%)} &
    32.221 \textcolor{gray!70}{\scriptsize (-0.57\%)} \\

    RALU & 30 & 11.03 & 2.34$\times$ & 1499.79 & 2.48$\times$ &
    0.9644 \textcolor{gray!70}{\scriptsize (-0.94\%)} &
    32.198 \textcolor{gray!70}{\scriptsize (-0.64\%)} \\
    
    \rowcolor{gray!20}
    $\textbf{Fresco}$  &  30 & \bf 9.20 & \textbf{2.81$\times$} & \bf1295.99  & \textbf{2.87$\times$} & 
    \bf 1.0527 \textcolor[HTML]{0f98b0}{\scriptsize \textbf{(+8.13\%)}} &
    \bf 32.521 \textcolor[HTML]{0f98b0}{\scriptsize \textbf{(+0.36\%)}} \\

    \midrule

    $36\%$ steps & 18 & 9.89 & 2.61$\times$ & 1339.02 & 2.77$\times$ &
    0.9553 \textcolor{gray!70}{\scriptsize (-1.88\%)} &
    32.114 \textcolor{gray!70}{\scriptsize (-0.89\%)} \\

    ToCa $(\mathcal{N}=6)$ & 50 & 13.16 & 1.96$\times$ & 924.30 & 4.02$\times$ &
    0.9702 \textcolor{gray!70}{\scriptsize (-0.35\%)} &
    32.083 \textcolor{gray!70}{\scriptsize (-0.99\%)} \\
    
    DuCa $(\mathcal{N}=5)$ & 50 & 8.18 & 3.15$\times$ & 978.76 & 3.80$\times$ &
    0.9855 \textcolor{gray!70}{\scriptsize (+1.22\%)} &
    32.241 \textcolor{gray!70}{\scriptsize (-0.50\%)} \\
    
    TeaCache $({l}=0.8)$  & 50 & 6.64 & 3.89$\times$ & 892.35 & 4.17$\times$ &
    0.8805 \textcolor{gray!70}{\scriptsize (-9.57\%)} &
    31.827 \textcolor{gray!70}{\scriptsize (-1.78\%)} \\

    TaylorSeer $(\mathcal{N}=4)$ & 50 & 9.24 & 2.80$\times$ & 967.91 & 3.84$\times$ &
    0.9857 \textcolor{gray!70}{\scriptsize (+1.25\%)} &
    32.413 \textcolor{gray!70}{\scriptsize (+0.03\%)} \\

    RALU & 18 & 6.34 & 4.07$\times$ & 904.98 & 4.11$\times$ &
    0.9481 \textcolor{gray!70}{\scriptsize (-2.62\%)} &
    32.274 \textcolor{gray!70}{\scriptsize (-0.40\%)} \\
     
    \rowcolor{gray!20}
    $\textbf{Fresco}$  &  18 & \bf 5.72 & \textbf{4.51$\times$} & \bf788.03 & \textbf{4.72$\times$} &
    \bf 1.0369 \textcolor[HTML]{0f98b0}{\scriptsize \textbf{(+6.50\%)}} &
    \bf 32.581 \textcolor[HTML]{0f98b0}{\scriptsize \textbf{(+0.55\%)}} \\

    \midrule

    ToCa $(\mathcal{N}=10)$ & 50 & 7.93 & 3.25$\times$ & 714.66 & 5.20$\times$ &
    0.7055 \textcolor{gray!70}{\scriptsize (-27.5\%)} &
    31.808 \textcolor{gray!70}{\scriptsize (-1.84\%)} \\

    DuCa $(\mathcal{N}=9)$ & 50 & 7.27 & 3.55$\times$ & 690.25 & 5.39$\times$ &
    0.8182 \textcolor{gray!70}{\scriptsize (-15.9\%)} &
    31.759 \textcolor{gray!70}{\scriptsize (-1.99\%)} \\
    
    TeaCache $({l}=1.6)$  & 50 & 3.78 & 6.82$\times$ & 520.54 & 7.15$\times$ &
    0.6423 \textcolor{gray!70}{\scriptsize (-34.0\%)} &
    31.656 \textcolor{gray!70}{\scriptsize (-2.31\%)} \\

    TaylorSeer $(\mathcal{N}=9)$ & 50 & 4.85 & 5.32$\times$ & 596.07 & 6.24$\times$ &
    0.8562 \textcolor{gray!70}{\scriptsize (-12.1\%)} &
    31.653 \textcolor{gray!70}{\scriptsize (-2.32\%)} \\

     FLUX.1-schnell &  8 & 4.21 & 6.12$\times$ & 595.12 & 6.25$\times$ &
    0.9097 \textcolor{gray!70}{\scriptsize (-6.56\%)} &
    \bf 33.837 \textcolor{gray!70}{\scriptsize (+4.42\%)} \\

    RALU & 10 & 3.73 & 6.92$\times$ & 540.62 & 6.88$\times$ &
    0.9289 \textcolor{gray!70}{\scriptsize (-4.59\%)} &
    32.113 \textcolor{gray!70}{\scriptsize (-0.90\%)} \\
    
    \rowcolor{gray!20}
    $\textbf{Fresco}$  & 11 &  3.43 &{7.52$\times$} &  486.85 & {7.64$\times$} &
    \bf 1.0090 \textcolor[HTML]{0f98b0}{\scriptsize \textbf{(+3.62\%)}} &
    32.370 \textcolor[HTML]{0f98b0}{\scriptsize \textbf{(-0.10\%)}} \\

    \rowcolor{gray!20}
    $\textbf{Fresco}$  & 9 & \bf 2.54 & \textbf{10.16$\times$} & \bf 362.17 & \textbf{10.27$\times$} &
    0.9825 \textcolor[HTML]{0f98b0}{\scriptsize \textbf{(+0.91\%)}} &
    32.426 \textcolor[HTML]{0f98b0}{\scriptsize \textbf{(+0.07\%)}} \\

    \bottomrule
\end{tabular}}
\label{table:FLUX}
\footnotesize
\end{table*}

\subsubsection{Progressive Upsampling}

To achieve fine-grained detail while avoiding redundant computation, Fresco selectively upsamples only semantically stable regions instead of the entire latent grid. During progressive sampling, we estimate the temporal variance of each token $\mathbf{z}_i^{(t)}$ across timesteps to evaluate its convergence state:
\begin{equation}
v_i = \mathrm{Var}_t(\mathbf{z}_i^{(t)}),
\end{equation}
where a smaller $v_i$ indicates stable semantic structure. A threshold $\tau$ is introduced to control the selection ratio, adaptively balancing the trade-off between efficiency and detail recovery. Tokens satisfying $v_i \leq \tau$ are considered converged and thus promoted early to higher resolution for detailed refinement, while those with larger variance remain at lower resolution to continue efficient denoising until stabilization. For tokens selected for refinement, we perform a spatial expansion upsampling. Each parent token $\mathbf{z}_{\text{parent}}\in\mathbb{R}^D$ is expanded into four child tokens using an orthogonal Hadamard transformation. Three independent Gaussian noise vectors $\epsilon_1,\epsilon_2,\epsilon_3\sim\mathcal{N}(0,I)$ are sampled, and the child tokens are obtained as
\begin{equation}
[\mathbf{z}_1,\mathbf{z}_2,\mathbf{z}_3,\mathbf{z}_4] = H_4 \cdot [\mathbf{z}_{\text{parent}},\,\epsilon_1,\,\epsilon_2,\,\epsilon_3],
\end{equation}
where $H_4$ is a $4\times4$ Hadamard matrix. This operation injects controlled orthogonal perturbations into each child token, which provides stochastic degrees of freedom for fine-scale texture in later diffusion process, while preserving the coarse semantic structure of the parent token.

\section{Experiments}

\begin{figure*}[t]
  \centering
  \includegraphics[trim=60 20 50 25, clip,width=\linewidth]{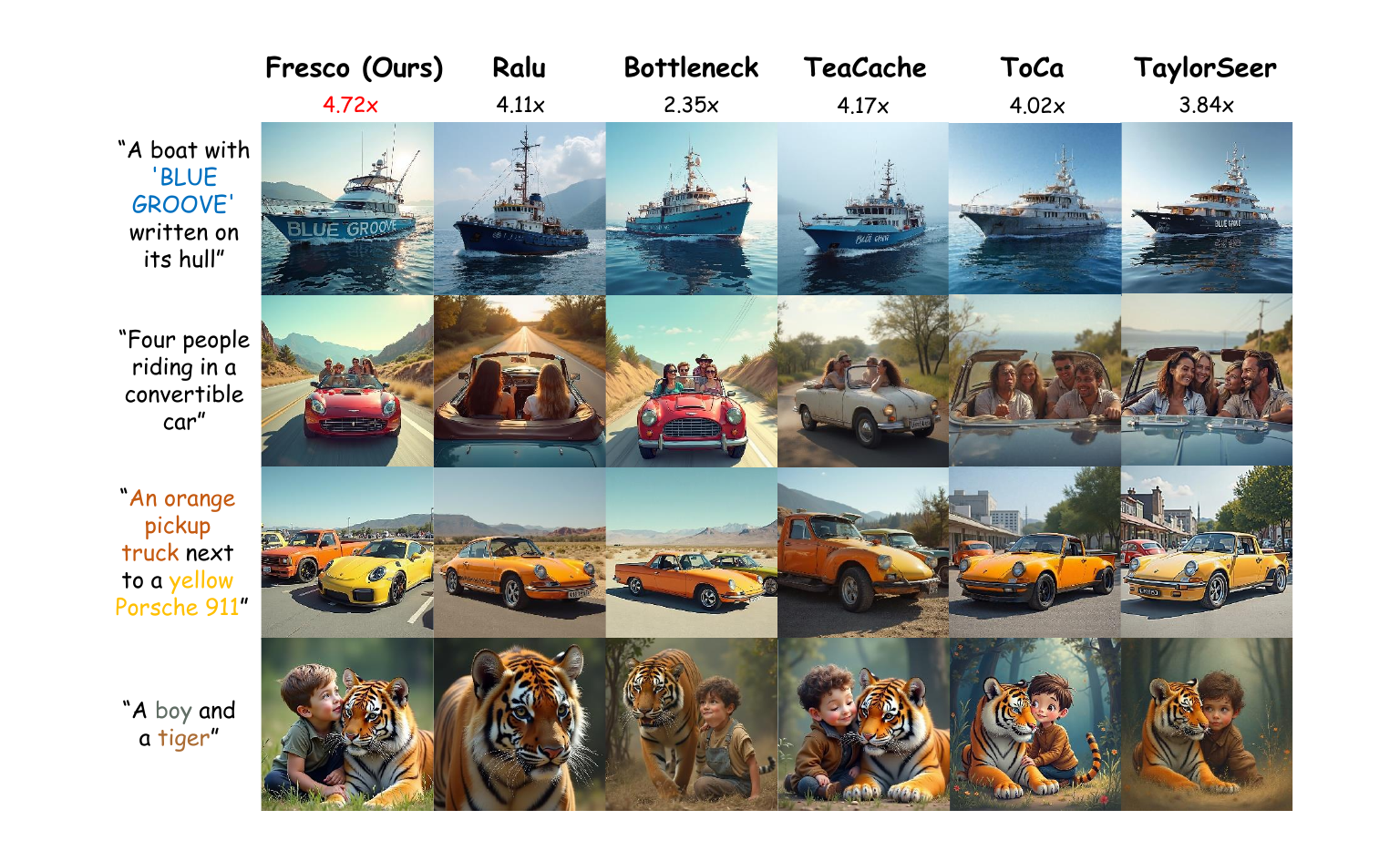}
  \caption{\textbf{Visualization of the image generated by different methods on FLUX.1-dev.} Fresco delivers the most realistic and semantically faithful results while achieving the fastest speed (4.72$\times$), outperforming all dynamic resolution and feature caching baselines.}
  \label{fig:flux}
\end{figure*}

\subsection{Experiment Settings}

\noindent\textbf{Model Configurations.} We conduct experiments on the text-to-image model {FLUX.1-dev}~\citep{flux2024} and the text-to-video model {HunyuanVideo}~\citep{sun_hunyuan-large_2024}. To further assess compatibility with other acceleration techniques, we additionally evaluate our method on quantized model FLUX.1-dev-torchao-int8~\citep{flux1dev_torchao_int8}, the model-distillation variant FLUX.1-lite-8B~\citep{flux1lite_8B}, and the step-distillation model FLUX.1-schnell~\citep{flux1_schnell}. These variants allow us to verify that our approach integrates seamlessly with other acceleration techniques, demonstrating broad compatibility across efficiency-oriented models.

\noindent\textbf{Evaluation and Metrics.} For text-to-image generation, we follow the DrawBench~\citep{saharia2022drawbench} protocol and evaluate all models on a fixed set of 200 prompts under 1024$\times$1024 resolution. We evaluate images using ImageReward~\citep{xu2023imagereward} and CLIP-IQA for photorealism, CLIP Score~\citep{hessel2021clipscore} for text–image alignment. For text-to-video generation, we evaluate HunyuanVideo on VBench~\citep{VBench}, which provides multi-dimensional human-aligned assessments of motion quality, visual appearance, and semantic consistency. Each generated video has a resolution of 720$\times$1280 and contains 125 frames. \emph{Please refer to Appendix for detail implementation.}

\begin{table*}[htb]
\centering
\caption{\textbf{Quantitative comparison of text-to-video generation} on HunyuanVideo.}

\setlength\tabcolsep{5.0pt} 
\small
\resizebox{\textwidth}{!}{
\begin{tabular}{l | c | c  c | c  c | c | c | c }
    \toprule
    \multirow{2}{*}{\centering \bf Method} & \multirow{2}{*}{\centering \bf NFE} &\multicolumn{4}{c|}{\bf Acceleration} & \multirow{2}{*}{\centering \bf  Quality Score $\uparrow$} &  \multirow{2}{*}{\centering \bf  Semantic Score $\uparrow$}  & \multirow{2}{*}{\centering \bf  Total Score $\uparrow$} \\
    \cline{3-6}
     &  & {\bf Latency(s) $\downarrow$} & {\bf Speed $\uparrow$} & {\bf FLOPs(T) $\downarrow$}  & {\bf Speed $\uparrow$} & & & \rule{0pt}{2ex}\\ 
    \midrule

  HunyuanVideo 
                           & 50  & 1887.12 & 1.00$\times$& {534440}   & {1.00$\times$} & 82.39 \textcolor{gray!70}{\scriptsize (+0.00\%)} & 71.04 \textcolor{gray!70}{\scriptsize (+0.00\%)} & 80.12 \textcolor{gray!70}{\scriptsize (+0.00\%)} \\ 
    \midrule
  
  {$22\%${ steps}}  & 11  & 414.75 & 4.55$\times$ & 116945   & {4.57$\times$} & 80.52 \textcolor{gray!70}{\scriptsize (-2.27\%)} & 69.01 \textcolor{gray!70}{\scriptsize (-2.86\%)} & 77.74 \textcolor{gray!70}{\scriptsize (-2.97\%)} \\

TeaCache
                           & 50 & 436.83 & 4.32$\times$ & 119294 & 4.48$\times$ & 81.12 \textcolor{gray!70}{\scriptsize (-1.54\%)} & 69.64 \textcolor{gray!70}{\scriptsize (-1.97\%)} & 78.96 \textcolor{gray!70}{\scriptsize (-1.45\%)} \\

AdaCache
                           & 50 & 720.27 & 2.62$\times$ & 200917 & 2.66$\times$ & 81.78 \textcolor{gray!70}{\scriptsize (-0.74\%)} & 70.12 \textcolor{gray!70}{\scriptsize (-1.29\%)} & 79.58 \textcolor{gray!70}{\scriptsize (-0.67\%)} \\

Bottleneck Sampling
                           & 30  & 792.85 & 2.38$\times$ & 204766  & 2.61$\times$ & 81.97 \textcolor{gray!70}{\scriptsize (-0.51\%)} & 69.55 \textcolor{gray!70}{\scriptsize (-2.10\%)} & 79.42 \textcolor{gray!70}{\scriptsize (-0.87\%)} \\

Jenga-ProRes
                           & 25  & 882.14 & 2.14$\times$ & 233379  & 2.29$\times$ & 81.80 \textcolor{gray!70}{\scriptsize (-0.72\%)} & 68.56 \textcolor{gray!70}{\scriptsize (-3.49\%)} & 79.16 \textcolor{gray!70}{\scriptsize (-1.20\%)} \\

\rowcolor{gray!20}
\textbf{Fresco}  
                           & 23  & 537.21 & {3.51$\times$} & 136685  & 3.91$\times$ & \bf 83.60 \textcolor[HTML]{0f98b0}{\scriptsize \textbf{(+1.47\%)}} & \bf 71.11 \textcolor[HTML]{0f98b0}{\scriptsize \textbf{(+0.10\%)}} & \bf 81.10 \textcolor[HTML]{0f98b0}{\scriptsize \textbf{(+1.22\%)}} \\

\rowcolor{gray!20}
\textbf{Fresco} 
                           &18   & \bf 403.23 & \textbf{4.68$\times$} & \bf 108626  & \bf {4.92$\times$} &  83.53 \textcolor[HTML]{0f98b0}{\scriptsize \textbf{(+1.38\%)}} &  69.65 \textcolor[HTML]{0f98b0}{\scriptsize \textbf{(-1.96\%)}} &  80.76 \textcolor[HTML]{0f98b0}{\scriptsize \textbf{(+0.80\%)}} \\

    \bottomrule
\end{tabular}}
\label{table:HunyuanVideo-Metrics}
\end{table*}

\begin{table*}[htbp]
\centering
\caption{\textbf{Quantitative comparison of text-to-image generation} with other accleration methods.}
\setlength\tabcolsep{7pt} 
\small
\resizebox{\textwidth}{!}{
\begin{tabular}{l | c | c  c | c  c | c | c }
    \toprule
    \rowcolor{white}
    \multirow{2}{*}{\centering \bf Method} & \multirow{2}{*}{\centering \bf NFE} & \multicolumn{4}{c|}{\bf Acceleration} & \multirow{2}{*} {\bf CLIP-IQA $\uparrow$} & \multirow{2}{*} {\bf CLIP Score $\uparrow$}\rule{0pt}{2ex} \\
    \cline{3-6}
    & & {\bf Latency(s) $\downarrow$} & {\bf Speed $\uparrow$} & {\bf FLOPs(T) $\downarrow$}  & {\bf Speed $\uparrow$} &  & \\

    \midrule
    {FLUX.1-dev}~\citep{flux2024} & 50 & 25.82 & 1.00$\times$ & 3719.50 & 1.00$\times$ & 0.9110 \textcolor{gray!70}{\scriptsize (+0.00\%)} & 32.404 \textcolor{gray!70}{\scriptsize (+0.00\%)} \\

    \midrule
    {Taylorseer}~\citep{liu2025reusingforecastingacceleratingdiffusion} & 50 & 9.24 & 2.79$\times$ & 967.91 & 3.84$\times$ & 0.8872 \textcolor{gray!70}{\scriptsize (-2.61\%)} & 32.413 \textcolor{gray!70}{\scriptsize (+0.03\%)} \\
    
    \rowcolor{gray!20}
    \textbf{Fresco}+Feature Caching  &  18 & \bf 2.89 & \textbf{8.93$\times$} & \bf 402.97  & \textbf{9.23$\times$} & \bf 0.9116 \textcolor[HTML]{0f98b0}{\scriptsize \textbf{(+0.07\%)}} & \bf 32.511 \textcolor[HTML]{0f98b0}{\scriptsize \textbf{(+0.33\%)}} \\
    
    \midrule

    {FLUX.1-lite-8B}~\citep{flux1lite_8B} & 28 & 9.18 & 2.81$\times$ & 1291.49 & 2.88$\times$ & 0.8938 \textcolor{gray!70}{\scriptsize (-1.88\%)} & \bf 32.135 \textcolor{gray!70}{\scriptsize (-0.83\%)} \\
    
    \rowcolor{gray!20}
    \textbf{Fresco}+Model Distillation  &  14 & \bf 2.55 & \textbf{10.12$\times$} & \bf 370.83  & \textbf{10.03$\times$} & \bf 0.9518 \textcolor[HTML]{0f98b0}{\scriptsize \textbf{(+4.48\%)}} &  31.426 \textcolor[HTML]{0f98b0}{\scriptsize \textbf{(-3.02\%)}} \\

    \midrule

    {FLUX.1-schnell}~\citep{flux1_schnell} & 8 & 4.21 & 6.12$\times$ & 595.12 & 6.25$\times$ & 0.8804 \textcolor{gray!70}{\scriptsize (-3.36\%)} & 33.837 \textcolor{gray!70}{\scriptsize (+4.42\%)} \\

    {FLUX.1-schnell}~\citep{flux1_schnell} & 4 & 2.77 & 9.32$\times$ & 283.56 & 13.11$\times$ & 0.8771 \textcolor{gray!70}{\scriptsize (-3.72\%)} & \bf 34.056 \textcolor{gray!70}{\scriptsize (+5.09\%)} \\
    
    \rowcolor{gray!20}
    \textbf{Fresco}+Step Distillation  &  8 & 3.22 & {8.01$\times$} & 371.95  &{10.53$\times$} & \bf 0.9541 \textcolor[HTML]{0f98b0}{\scriptsize \textbf{(+4.73\%)}} &  32.909 \textcolor[HTML]{0f98b0}{\scriptsize \textbf{(+1.56\%)}} \\
    
    \rowcolor{gray!20}
    \textbf{Fresco}+Step Distillation  &  4 & \bf 1.87 & \textbf{13.08$\times$} & \bf177.25  & \textbf{22.10$\times$} & 0.8693 \textcolor[HTML]{0f98b0}{\scriptsize \textbf{(-4.58\%)}} &  32.563 \textcolor[HTML]{0f98b0}{\scriptsize \textbf{(+0.49\%)}} \\

    \midrule
    {FLUX.1-dev-int8}~\citep{flux1dev_torchao_int8} & 50 & 14.03 & 1.83$\times$ & 1888.07 & 1.97$\times$ & 0.9082 \textcolor{gray!70}{\scriptsize (-0.31\%)} & 32.404 \textcolor{gray!70}{\scriptsize (+0.00\%)} \\
    
    \rowcolor{gray!20}
    \textbf{Fresco}+Quantization  &  30 &  5.25 & {4.91$\times$} &  674.99  & 5.51$\times$ & \bf 0.9097 \textcolor[HTML]{0f98b0}{\scriptsize \textbf{(-0.14\%)}} &  31.442 \textcolor[HTML]{0f98b0}{\scriptsize \textbf{(-2.97\%)}} \\

    \rowcolor{gray!20}
    \textbf{Fresco}+Quantization  &  18 & \bf 3.12 & \textbf{8.30$\times$} & \bf 400.37  & \textbf{9.29$\times$} &  0.8913 \textcolor[HTML]{0f98b0}{\scriptsize \textbf{(-2.16\%)}} & \bf 32.526 \textcolor[HTML]{0f98b0}{\scriptsize \textbf{(+0.38\%)}} \\

    \bottomrule
\end{tabular}}
\label{table:distill}
\footnotesize
\end{table*}

\subsection{Results on Text-to-Image Generation}

As shown in Table~\ref{table:FLUX}, Fresco consistently outperforms existing acceleration methods on FLUX.1-dev. 
At moderate acceleration, it achieves an ImageReward of \textbf{1.0527} and CLIP score of 32.521 with a 2.87$\times$ FLOPs reduction, surpassing TeaCache (0.9449) and Bottleneck Sampling (0.9739). Under stronger compression, Fresco maintains high quality (ImageReward 1.0369, CLIP 32.581) with a 4.72$\times$ speedup, outperforming TaylorSeer (0.9857) and RALU (0.9481). Even under extreme acceleration, Fresco retains strong fidelity (ImageReward 1.0090) at 7.64$\times$ speedup. At the most aggressive setting, Fresco still reaches an ImageReward of \textbf{0.9825} which is even higher than the original model and the step-distillation model FLUX.1-schnell, with \textbf{10.27$\times$} speedup. Visualization in Fig.~\ref{fig:flux} further demonstrate Fresco’s superior efficiency while preserving image quality.

\begin{figure*}[t]
  \centering
  \includegraphics[trim=130 50 150 20, clip,width=\linewidth]{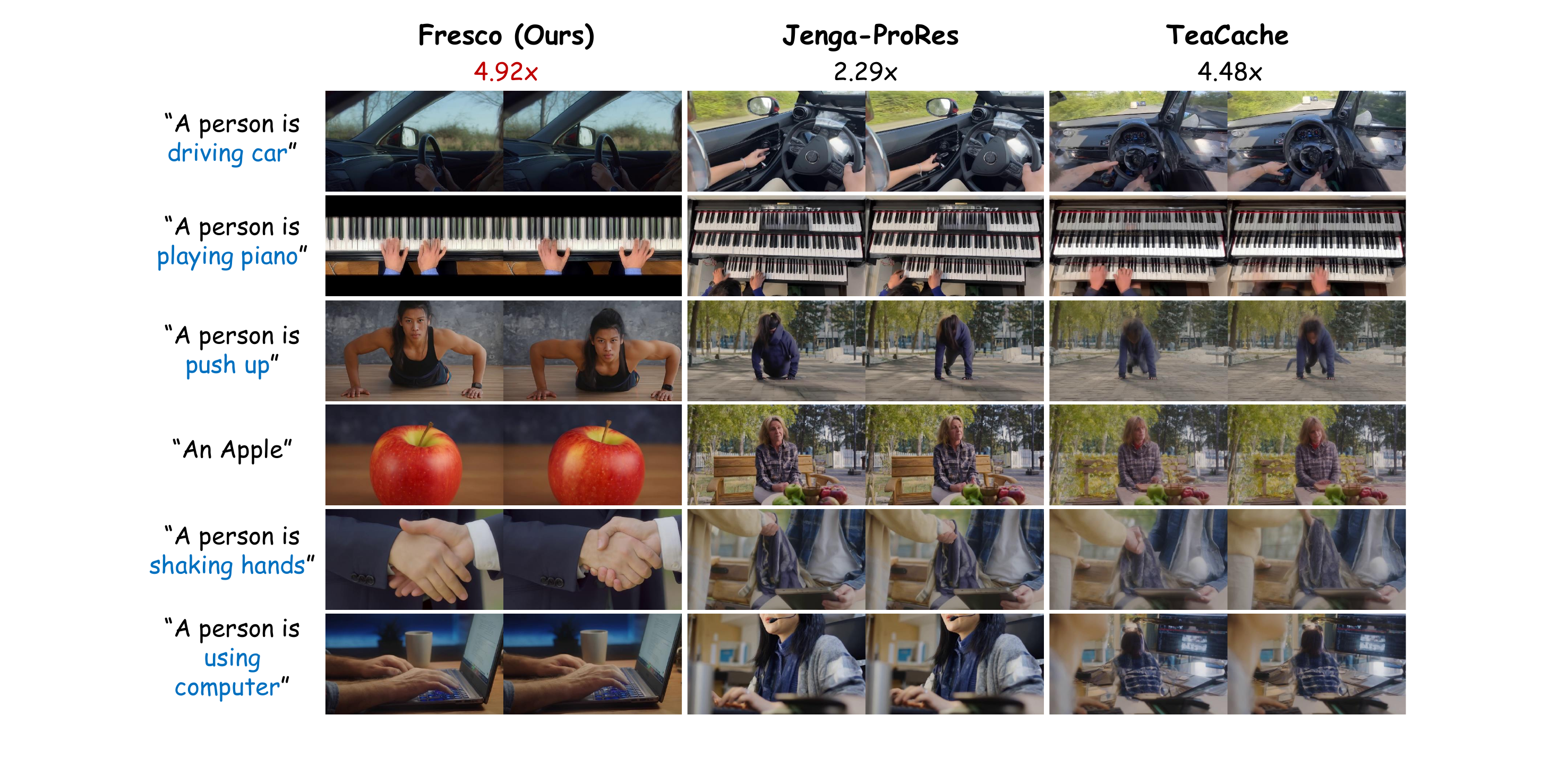}
  \caption{\textbf{Visualization of different acceleration methods on HunyuanVideo.} Fresco achieves the best visual quality, semantic accuracy, and highest speedup ratio 4.92$\times$ among all methods, demonstrating strong generalization ability on video generation models.}
  \label{fig:video}
\end{figure*}

\subsection{Results on Text-to-Video Generation}
Table~\ref{table:HunyuanVideo-Metrics} shows that Fresco consistently achieves strong results on HunyuanVideo across different acceleration settings. At moderate compression, Fresco achieves the highest total score of \textbf{81.10} with a 3.91$\times$ FLOPs reduction, surpassing TeaCache (78.96) and Jenga-ProRse (79.16) while improving quality and semantic scores. Under stronger acceleration, Fresco maintains competitive performance with a total score of 80.76 at a \textbf{4.92$\times$} speedup, outperforming all baselines of comparable efficiency. The visual comparison on Fig.~\ref{fig:video} further underscores the effectiveness of globally consistent noise and progressive refinement for dynamic spatiotemporal generation with precise semantics.

\subsection{Compatibility with Acceleration Methods}

To further validate generality, we evaluate Fresco in combination with several mainstream acceleration techniques, including feature caching, quantization, and distillation. As shown in Table~\ref{table:distill}, Fresco consistently enhances performance across all settings without introducing additional degradation. When integrated with feature caching method TaylorSeer, Fresco achieves a 9.23$\times$ acceleration and improves CLIP-IQA from 0.8872 to 0.9116. On model-distillation variant, Fresco delivers substantial gains, reaching a 2.87$\times$ speed-up with improvement in CLIP-IQA. Notably, Fresco combined with step-distillation achieves up to \textbf{22.10$\times$} acceleration while maintaining comparable quality (0.8693 CLIP-IQA, 32.563 CLIP). Integration with INT8 quantization provides a 9.29$\times$ speedup and stable perceptual quality. These results demonstrate that Fresco is highly compatible with diverse efficiency-oriented methods, serving as a complementary framework that amplifies their acceleration benefits while preserving generation fidelity.

\begin{table}[ht]
    \centering
    \caption{\textbf{Ablation study on token selection methods}}
    \setlength\tabcolsep{4pt}
    \small
    
    \resizebox{0.48\textwidth}{!}{\begin{tabular}{l | c  c | c c}
    \toprule
    \bf Method
      & \bf Latency(s) $\downarrow$

      & \bf Speed $\uparrow$
      & \bf ImageReward $\uparrow$
      & \bf ClipScore $\uparrow$ \\
    \midrule
    {Random}
      & 5.26  & 4.90$\times$ & 0.9143  & 31.277 \\
    {Edge detection}
      & 5.77  & 4.47$\times$ & 0.9482  & 32.316 \\
    {Attention score}
      & 6.13  & 4.21$\times$ & 0.9849  & 32.324\\
    \rowcolor{gray!20}
    \textbf{Variance}
      & 5.72 & 4.51$\times$ & \textbf{1.0369}  & \textbf{32.581} \\
    \bottomrule
    \end{tabular}
    }
    \label{table:variance}
\end{table}

\begin{table}[ht]
    \centering
    \caption{\textbf{Acceleration on different resolutions}}
    \setlength\tabcolsep{4pt}
    \small
    
    \resizebox{0.48\textwidth}{!}{\begin{tabular}{l | c | c c | c }
    \toprule
    \bf Resolution
    & \bf Method 
      & \bf Latency(s) $\downarrow$
        
      & \bf Speed $\uparrow$
      & \bf ImageReward $\uparrow$
 \\
    \midrule
    \multirow{2}{*}{\centering \bf 1024$\times$1024} & FLUX
      & 25.82  & 1.00$\times$ & 0.9736 \\
& Fresco & \bf 5.72  & \bf4.51$\times$ & \bf1.0369 \\
          \midrule
      
    \multirow{2}{*}{\centering \bf 1440$\times$1440} & FLUX
      & 52.42  & 1.00$\times$ & 0.8873 \\

& Fresco & \bf9.58  & \bf5.47$\times$ & \bf0.9206 \\
          \midrule
      
    \multirow{2}{*}{\centering \bf 2048$\times$2048} & FLUX
      & 120.43  & 1.00$\times$ & 0.8499 \\

& Fresco & \bf21.20  & \bf5.68$\times$ & \bf0.9348 \\
    \bottomrule
    \end{tabular}
    }
    \label{table:resolution}
\end{table}

\subsection{Ablation Study}

We compare four strategies for selecting tokens to upsample early: random, edge detection, attention score, and variance. As shown in Table~\ref{table:variance}, variance achieves the best balance, offering a 4.51$\times$ speedup, which is only slightly slower than random, while significantly outperforming in generation quality. This indicates that variance effectively identifies semantically stable tokens, enabling precise and efficient refinement.

Table~\ref{table:resolution} shows results under different resolutions. Our method consistently improves speed and quality over the original FLUX. The speedup ratio increases with resolution, from 4.51$\times$ to 5.68$\times$, due to the greater redundancy in high-resolution inputs. Meanwhile, ImageReward also improves, demonstrating our method’s scalability and fidelity.

\begin{figure}[htbp]
  \centering
  \includegraphics[trim=55 30 55 40, clip,width=1\linewidth]{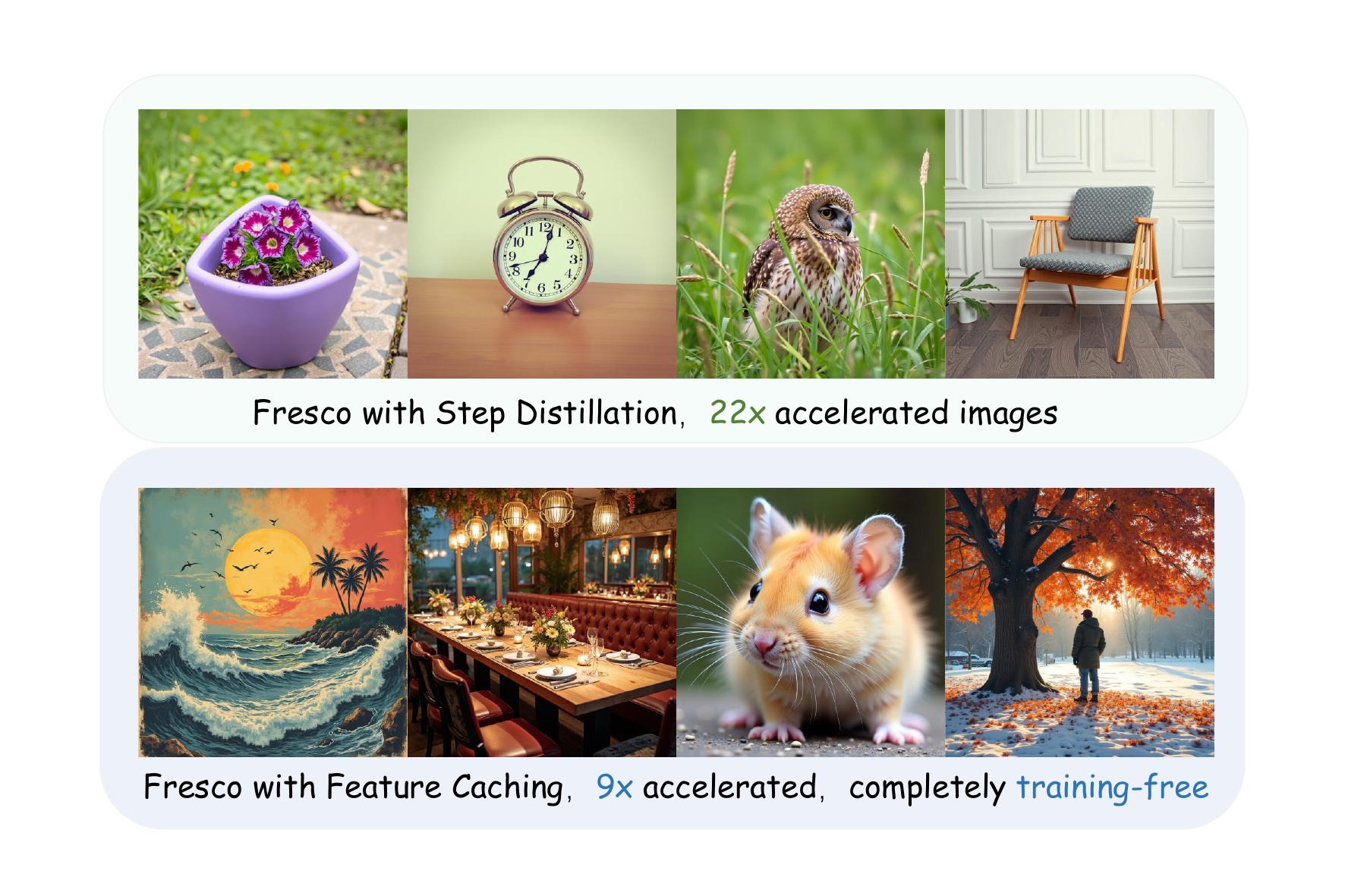}
  \caption{\textbf{Compatibility with other acceleration methods.} Fresco achieves 22$\times$ speedup when combined with step distillation and 9$\times$ training-free acceleration when paired with feature caching, while preserving high visual fidelity.}

  \label{fig:distill}
\end{figure}

\begin{figure}[htbp]
  \centering
  \includegraphics[trim=55 20 40 60, clip,width=1\linewidth]{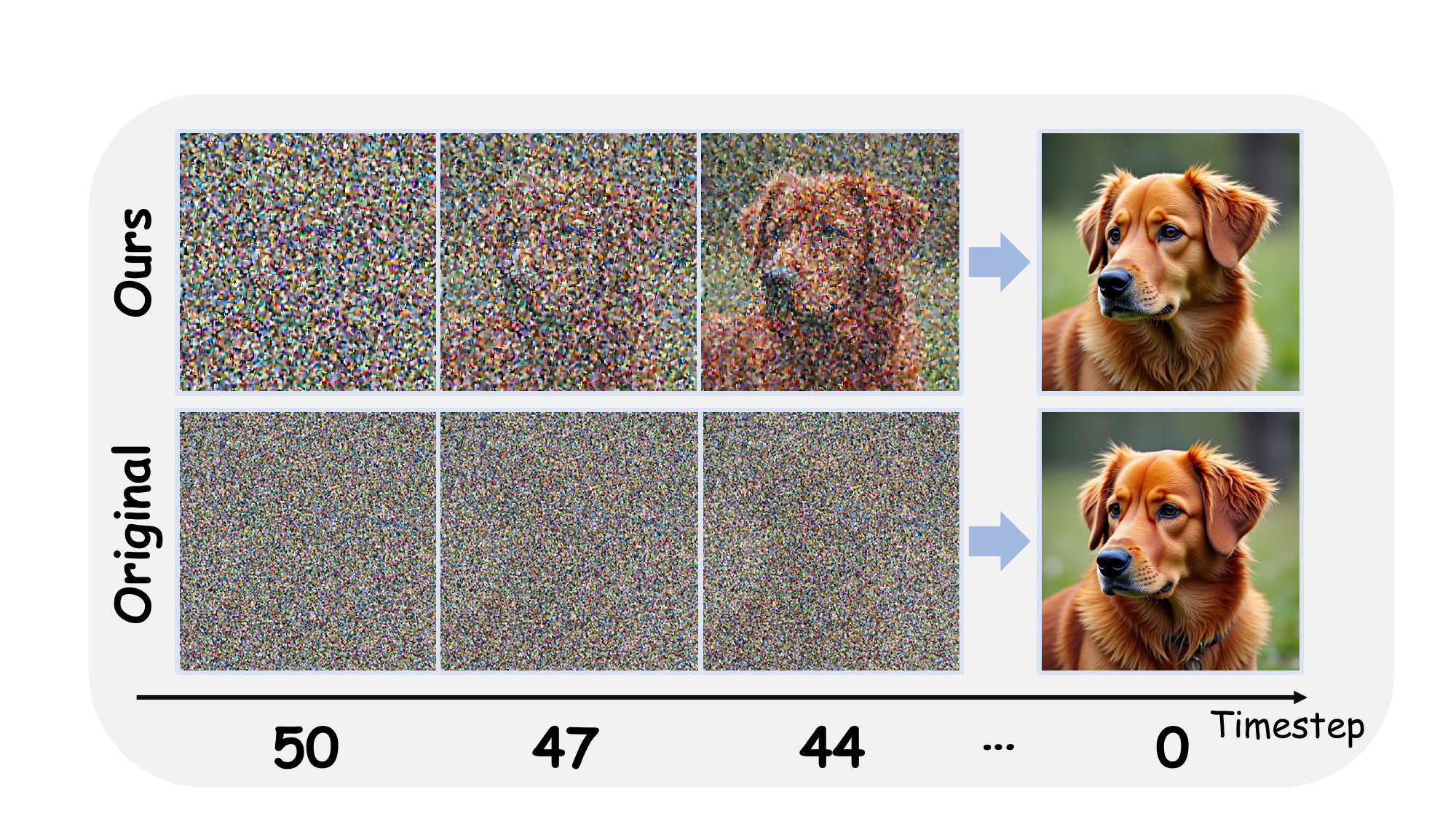}
  \caption{\textbf{Fast convergence with fewer steps.} Fresco drafts the global structure within the first few low-resolution steps (50→44), while the original model remains dominated by noise.}

  \label{fig:discussion}
\end{figure}

\section{Discussion}

Fresco is particularly compatible with feature caching, especially recent forecast-based caching methods. Under the unified noise injection and variance-guided upsampling constraints, Fresco produces smoother feature trajectories across time. This temporal stability enhances the predictability of intermediate features, making Fresco a natural fit for caching methods like Taylorseer~\citep{liu2025reusingforecastingacceleratingdiffusion} and Foca~\citep{zheng2025forecastcalibratefeaturecaching}. A natural and intuitive way to combine Fresco with feature caching is to adapt caching intensity by resolution: in early low-resolution stages, each step is inexpensive, so fewer caches and more computations are preferred; in later high-resolution stages, where computation becomes costly, more caching and fewer computations are applied. This complementary design allows Fresco and feature caching to work synergistically as shown in Fig.~\ref{fig:distill}, achieving up to 9$\times$ acceleration in a fully training-free manner.

Additionally, current diffusion transformer models often exhibit degraded performance when generating high-resolution images. As shown in Table~\ref{table:resolution}, the generation quality of FLUX drops noticeably at resolutions of 1440$\times$1440 and above. This degradation typically manifests as overly local or fragmented outputs, primarily due to the fact that most diffusion transformer models are trained on low-resolution images. Fresco can alleviate this issue by constructing the image in a coarse-to-fine manner, beginning the denoising process at a lower resolution where the model performs best. For instance, to generate a 2048$\times$2048 image on FLUX, Fresco downsamples to 1024$\times$1024 and performs early-stage sampling in this resolution, which is precisely the resolution sweet spot for FLUX, thus preserving global structure and improving output fidelity at high resolutions.

Beyond reducing computational cost in early stages, we observe that Fresco also exhibits a notable degree of step-reduction capability. In many samples, Fresco constructs the global scene layout within the first 7-8 low-resolution steps, producing a coherent coarse sketch long before high-resolution refinement begins. In contrast, the original full-resolution sampler often struggles to form any meaningful structure with the same number of steps, as shown in Fig.~\ref{fig:discussion}. This might suggest that diffusion dynamics at reduced spatial resolution converges more rapidly, allowing the model to establish global geometry and semantic layout with significantly fewer iterations. By shifting early denoising into this fast-converging regime, Fresco not only reduces per-step cost but also reduces the number of sampling steps.

\section{Conclusion}

We presented Fresco, a training-free coarse-to-fine sampling framework that unifies low-resolution efficiency with high-resolution fidelity for diffusion transformers. By introducing a token-encoded unified noise field and variance-guided progressive upsampling, Fresco preserves semantic continuity across stages and allocates computation only to stable regions. This design enables substantial acceleration—up to 10× on FLUX and 5× on HunyuanVideo—while maintaining high image and video quality, and integrates seamlessly with distillation, quantization, and feature caching for even larger gains. Fresco offers a simple, general, and scalable paradigm for efficient high-resolution generation.

{
    \small
   \bibliographystyle{ieeenat_fullname}
    \bibliography{main}
}


\end{document}